\documentclass[preprint,12pt]{elsarticle}

\usepackage{amssymb}
\usepackage{verbatim}
\usepackage{float}
\usepackage{epstopdf}
\usepackage{algorithm,algcompatible,amsmath}
\usepackage{gensymb}
\usepackage{booktabs}
\usepackage{amssymb,verbatim,epstopdf,algorithm,algcompatible,amsmath,caption,subcaption,algcompatible}
\usepackage{xcolor}
\usepackage{subcaption}
\usepackage{algcompatible}
\usepackage[flushleft]{threeparttable}

\newcommand\mynotes[1]{\textcolor{red}{#1}}
\newcommand\tgknotes[1]{\textcolor{green}{#1}}
\renewcommand\mynotes[1]{} 
\renewcommand\tgknotes[1]{} 

\journal{Adaptive Behavior}

\begin{document}

\begin{frontmatter}



\title{Self-Organizing Maps for Storage and Transfer of Knowledge in Reinforcement Learning\footnote{This work in an extension of Karimpanal \& Bouffanais~\citep{karimpanalself}, which was presented at the Adaptive Learning Agents (ALA) workshop in Stockholm, Sweden in July, 2018  }}


\author{Thommen George Karimpanal, Roland Bouffanais}
\ead{thommen\_george@mymail.sutd.edu.sg, bouffanais@sutd.edu.sg}
\address{Singapore University of Technology and Design, 8 Somapah Road, Singapore 487372}

\begin{abstract}
The idea of reusing or transferring information from previously learned tasks (source tasks) for the learning of new tasks (target tasks) has the potential to significantly improve the sample efficiency of a reinforcement learning agent. 
In this work, we describe a novel approach for reusing previously acquired knowledge by using it to guide the exploration of an agent while it learns new tasks. In order to do so, we employ a variant of the \textit{growing self-organizing map} algorithm, which is trained using a measure of similarity that is defined directly in the space of the vectorized representations of the value functions. 
In addition to enabling transfer across tasks, the resulting map is simultaneously used to enable the efficient storage of previously acquired task knowledge in an adaptive and scalable manner. We empirically validate our approach in a simulated navigation environment, and also demonstrate its utility through simple experiments using a mobile micro-robotics platform. In addition, we demonstrate the scalability of this approach, and analytically examine its relation to the proposed network growth mechanism. Further, we briefly discuss some of the possible improvements and extensions to this approach, as well as its relevance to real world scenarios in the context of continual learning.
\end{abstract}


\end{frontmatter}

\section{Introduction}
The use of off-policy algorithms~\citep{geist2014off} in reinforcement learning (RL)~\citep{sutton2011reinforcement} has enabled the learning of multiple tasks in parallel. This is particularly useful for agents operating in the real world, where a number of tasks are likely to be encountered, and may be required to be learned~\citep{sutton2011horde,white2012scaling}. As more and more tasks are learned through agent-environment interactions, an ideal agent should be able to efficiently store and extract meaningful information from this accumulated knowledge and use it to accelerate its learning on new, related tasks. 
This is an active area of research in RL, referred to as transfer learning \citep{taylor_transfer_2009}. 

Formally, transfer learning is an approach to improve learning performance on a new `target' task $M_{T}$, using accumulated knowledge from a set of `source' tasks, $M_{S}=\{M_{s_{1}},..M_{s_{i}},..M_{s_{n}}\}$. Here, each task $M$ is a \textit{Markov Decision Process (MDP)}~\citep{Puterman:1994:MDP:528623}, such that $M=\{\mathcal S,\mathcal A,\mathcal T,\mathcal R\}$, where $\mathcal S$ is the state space, $\mathcal A$ is the action space, $\mathcal T$ is the transition function, and $\mathcal R$ is the reward function. As in some recent works \citep{barreto2017successor,laroche2017transfer}, we address the relatively simple case where tasks vary only in the reward function $\mathcal R$, while $\mathcal S, \mathcal A$ and $\mathcal T$ remain fixed across the tasks. 
For knowledge transfer to be effective, source tasks need to be selected appropriately. Reusing knowledge from an inappropriately selected source task could lead to negative transfer \citep{Lazaric2012,taylor_transfer_2009}, which is detrimental to the learning of the target task. In order to avoid such problems and ensure a beneficial transfer, a number of MDP similarity metrics \citep{ferns2004metrics,carroll2005task} have been proposed.  
However, it has been shown that the optimal MDP similarity metric to be used is dependent on the transfer mechanism employed~\citep{carroll2005task}. In addition, for an agent interacting with its environment, value functions pertaining to numerous tasks may be learned over a period of time. Some of these tasks may be very similar to each other, which could result in considerable redundancy in the stored value function information. Traditional transfer mechanisms are generally not designed to handle situations involving a large number of source tasks, which a real world agent could possibly encounter. From a continual learning perspective, a suitable mechanism is needed to enable the storage of such information in a scalable manner. 

In this work, we represent value functions ($Q$-values) using linear function approximation~\citep{sutton2011reinforcement},
and the knowledge of a particular task is assumed to be contained in the learned weights associated with the corresponding value ($Q$-) function. We define a cosine similarity metric within this value function weight space, and use this as a basis for maintaining a scalable knowledge base, while simultaneously using it to perform knowledge transfer across tasks. 
This is achieved using a variant of the growing self organizing map (GSOM) \citep{alahakoon2000dynamic}. The inputs to this GSOM algorithm consist of 
the value function weights of newly learned tasks, along with any previously learned knowledge that was stored in the nodes of the self-organizing map (SOM). During the GSOM training process, the winning node is selected based on the cosine similarity metric mentioned above. 
As the agent interacts with its environment and learns the value function weights corresponding to new tasks, this new information is incorporated into the map, which 
evolves by growing (if needed) to a suitable size in order to sufficiently represent all of the agent's gathered knowledge. 
Each element/node of the resulting map is a variant of the input value function weights (knowledge of previously learned tasks). These variants are treated as solutions to arbitrary source tasks, each of which is related to some degree to one of the previously learned tasks.  It is worth mentioning that the aim of storing knowledge in this manner is not to retain the exact value function information corresponding to all previously learned tasks, but to maintain a compressed and scalable knowledge base that can approximate the value function weights of previously learned tasks. Such approximations may be necessary in applications such as mobile robotics, where on-board memory is typically limited.

While learning a new target task, this knowledge base is used to identify the most relevant source task, based on the same similarity metric. 
The value function associated with this task is then greedily exploited to provide the agent with action advice to 
guide it towards achieving the target task. Due to the random initialization of the weights, the agent's initial estimates of the target task value function weights is expected to be poor. Consequently, it is unlikely that appropriate tasks would be selected for transfer at this stage. However, as the agent gathers more experience through its interactions with the environment, these estimates improve, which consequently leads to improvements in the estimates of the similarities between the target and source tasks. As a result, the agent becomes more likely to receive relevant action advice from a closely related source task. This action advice can be adopted, for instance, on an $\epsilon$-greedy basis, essentially substituting
the agent's exploration strategy. In this manner, the knowledge of source tasks can be used to merely guide the agent's exploratory behavior, thereby minimizing the risk of negative transfer which could have otherwise occurred, especially if value functions or representations were directly transferred between the tasks. Specifically, unlike direct transfer approaches, our approach only biases the agent's exploration strategy, and consequently, poor transfers are not catastrophic, and are relatively easier to withstand.

\begin{figure}[h]
\centering
\includegraphics[width=0.8\linewidth]{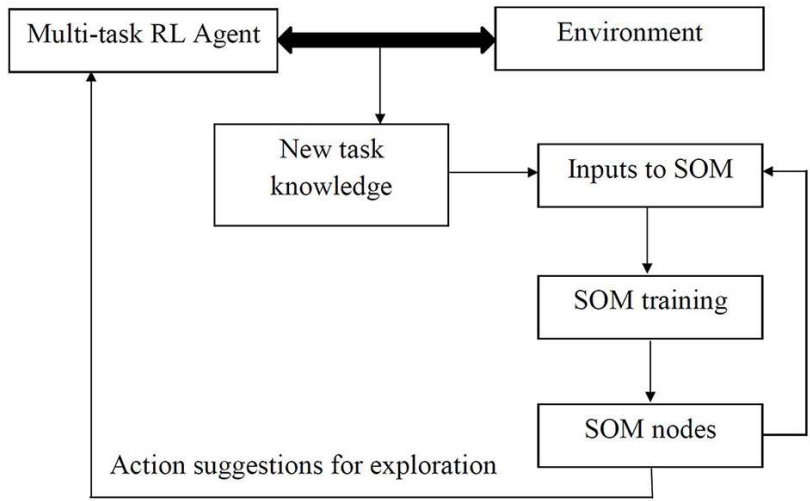}
\caption{The overall structure of the proposed SOM based knowledge storage and transfer approach.} 
\label{gen_struct}
\end{figure}

Hence, apart from maintaining an adaptive knowledge base of value function weights related to learned tasks, the proposed approach aims to leverage this knowledge base to make informed exploration decisions, which could lead to faster learning of target tasks. This could be especially useful in real world scenarios where factors such as learning speed and sample efficiency are critical, and several new tasks may need to be learned continuously, as and when they are encountered. The overall structure of the proposed methodology is depicted in Fig. \ref{gen_struct}.

\section{Related Work}
\label{relatedwork}
The sample efficiency of RL algorithms is one of the most critical aspects that determines the feasibility of its deployment in real world applications. Transfer learning is one of the mechanisms through which this issue can be addressed. Consequently, numerous techniques have been proposed \citep{Lazaric2012,taylor_transfer_2009,zhan2015online} to efficiently reuse the knowledge of learned tasks. A number of these \citep{carroll2005task,ammar2014automated,song2016measuring} rely on a measure of similarity between MDPs in order to choose an appropriate source task to transfer from. However, this can be problematic, as no such universal metric exists \citep{carroll2005task}, and some of the useful ones may be computationally expensive \citep{ammar2014automated}. In the present work, the similarity metric used is computationally inexpensive, and the degree of similarity between two tasks is based solely on the value function weights associated with them. The use of such a similarity metric, however, is restricted to cases where the MDPs vary only in their reward functions. Although some recent approaches such as the one described by Gupta et al. \citep{gupta2017learning} address the general case without such restrictions, it makes strong assumptions regarding the existence of structural similarities in the reward functions of the target and source tasks. This approach primarily focuses on the transfer between agents having different state-action spaces and transition dynamics. In addition, it is not designed to handle multiple tasks, and cannot automatically select appropriate source tasks.

In the approach we describe here, once an appropriate source task is identified, its value functions are used solely to extract action advice, which is used to guide the exploration of the agent. Similar approaches to transfer learning using action advice have been reported in Torrey et al. \citep{torrey2013teaching}, Zhan et al. \citep{zhan2015online} and Zimmer et al. \citep{zimmer2014teacher} which adopt a teacher-student framework for RL. However, these works assume that an effective policy for a particular target task is already accessible to the teacher, which is not the case in the present work. 

SOM-based approaches have previously been used in RL for a number of applications such as improving learning speed \citep{Tateyama2004}, representation in continuous state-action domains \citep{SMITH20021107,MONTAZERI20111069}, etc. In the context of scaling task knowledge for continual learning \citep{ring1994continual}, Ring et al. \citep{ring2011two} described a modular approach to assimilate the knowledge of complex tasks using a training process that closely resembles SOM. In this approach, a complex task is decomposed into a number of simple modules, such that modules close to each other correspond to similar agent behaviors. 
Teng et al. \citep{teng2015self} proposed a SOM-based approach to integrate domain knowledge and RL, with the aim of developing agents that can continuously expand their knowledge in real time, through their interactions with the environment. These ideas of knowledge assimilation are also reflected in the present work, although we also aim to reuse this knowledge to aid the learning of other related tasks. 

The transfer mechanism described here is inherently tied to the SOM-based approach for maintaining the knowledge of learned tasks. Apart from SOM, other clustering approaches \citep{thrun1998clustering,liu2012transfer,carroll2005task} have also been applied to achieve transfer learning in RL. In one of the earliest notable approaches to transfer learning, Thrun et al. \citep{thrun1998clustering} described a methodology for transfer learning by clustering learning tasks using a nearest neighbor clustering approach. Task similarity was determined using a task transfer matrix, which helped localize the appropriate task cluster to transfer from. 

More recent methods, such as the approach of \emph{Universal Value Function Approximators}~\citep{schaul2015universal} attempt to achieve transfer across tasks by learning a unified value function approximator that generalizes over states as well as goals. However, due to the fact that the underlying structure in the state-goal space may be highly complex, such an approach would, in most cases, be dependent on computationally inefficient function approximators such as deep neural networks, which may be infeasible to train in many real world scenarios. Our approach, on the other hand, is applicable to a range of value function representation schemes (linear function approximation, tabular etc.,), and allows value functions to be learned using any standard off-policy method. The structure of the goal space is extracted separately, using SOMs.

Perhaps the most similar work is the \emph{Probabilistic Policy Reuse (PPR)} algorithm~\citep{fernandez2013learning}, in which previously learned policies are used to bias the exploratory actions of the agent when it learns a new task. In addition to applying this exploration bias, a library of policies is also maintained, based on the similarities in their average discounted returns per episode. These `core' policies are considered to be representative of the domain under consideration. Although the present work shares a very similar exploration strategy to the one used in PPR, the manner in which policies are chosen to provide exploratory action advice varies considerably. We hypothesize that the non-linear basis function in SOMs would allow for the domain structure to be extracted more accurately than the average return basis used in PPR. In addition, with the use of SOMs, different policies or value functions (and hence, different agent behaviors) can be mapped in relation to each other, and can be visually represented.

Apart from PPR, the recent `Actor-mimic' \citep{parisotto2015actor} approach also performs transfer using action advice. In this approach, useful behaviors of a set of expert policy networks are compressed into a single multi-task network, which is then used to provide action advice in an $\epsilon -$greedy manner. The authors also report the problem of dramatically varying ranges of the value function across different tasks, which is resolved by using a Boltzmann distribution function. In the present work, the use of the cosine similarity metric resolves this issue and ensures that the similarity measure between tasks is bounded. Cosine similarity measures have previously been used in machine learning applications \citep{huang2012learning,chunjie2017cosine}, but to the best of our knowledge, it has not been used as a basis for task similarity or transfer in reinforcement learning. Apart from being able to handle tasks with vastly different value functions, the use of such a similarity metric also shields against negative transfer to a certain extent, as it provides a basis for the appropriate selection of source tasks. In addition to this, the actor-mimic and other approaches ignore the issues of knowledge redundancy and scalable storage, both of which are explicitly addressed in the proposed SOM based approach.


\section{Methodology}
\label{method}
In this work, we present an approach that enables the reuse of knowledge from previously learned tasks to aid the learning of a new task. 
Our approach consists of two fundamental mechanisms: (a) the accumulation of learned value function weights into a knowledge base in a scalable manner, and (b) the use of this knowledge base to guide the agent during the learning of the target task. 
The basis for these mechanisms is centered around the task similarity metric we propose here. We consider two tasks to be similar based on the cosine similarity between their corresponding learned value function weight vectors. For instance, the cosine similarity $c_{{w_{1}},{w_{2}}}$ between two non-zero weight vectors $\vec{w_{1}}$ and $\vec{w_{2}}$ is given by:
\begin{equation}
\label{eqn1}
 c_{{w_{1}},{w_{2}}}=\vec{w_{1}}.\vec{w_{2}}/|\vec{w_{1}}||\vec{w_{2}}|.
\end{equation}
The key idea is that two tasks are more likely to be similar to each other if they have similar feature weightings. Using such a similarity metric has certain advantages, such as boundedness and the ability to handle weight vectors with largely different magnitudes. During the construction of the scalable knowledge base, the mentioned similarity metric (Eq.~\eqref{eqn1}) is used as a basis for training the self-organizing map. Once this map has been constructed, the cosine similarity is again used as a basis for selecting an appropriate source task weight vector to guide the exploratory behavior of the agent while it learns a new task. Initially, owing to poor estimates of the value function weights of the new task, the selected source task may not be appropriate. However, as these estimates improve, more appropriate source tasks are identified and the corresponding action advice becomes more likely to be relevant to the task at hand. We now describe these mechanisms in detail.

\subsection{Knowledge Storage Using Self-Organizing Map}
\label{SOM_approach}

A SOM~\citep{kohonen1998self} is a type of unsupervised neural network used to produce a low-dimensional representation of its high-dimensional training samples. Typically, a SOM is represented as a two- or three-dimensional grid of nodes. Each node of the SOM is initialized to be a randomly generated weight vector of the same dimensions as the input vector. During the SOM training process, an input is presented to the network, and the node that is most similar to this input is selected to be the `winner'. The winning node is then updated towards the input vector under consideration.
Other nodes in the neighborhood are also influenced in a similar manner, but as a function of their topological distances to the winner. 
The final layout of a trained map is such that adjacent nodes have a greater degree of similarity to each other in comparison to nodes that are far apart. In this way, the SOM extracts the latent structure of the input space. 

For our purposes, the knowledge of an RL task is assumed to be contained in its associated value function weights, which may be learned using a number of approaches \citep{sutton2011reinforcement}. A na\"ive approach to storing knowledge associated with a number of tasks is to explicitly store the value function weights of these tasks. Apart from the scalability issue associated with such an approach, if several of these tasks are very similar or nearly identical to each other, it could introduce a high degree of redundancy in the knowledge stored. A more generalized approach to knowledge storage would be to store the characteristic features of the weight vectors associated with the learned tasks. The ability of the SOM to extract these features in an unsupervised manner makes it an attractive choice for the proposed knowledge storage mechanism.

In our approach, a rectangular SOM topology is used, and the inputs to the SOM are learned value function weights of previously encountered/learned tasks (input tasks).
The hypothesis is that after training, the weight vectors associated with each node in the SOM have varying degrees of similarity to the input vectors, and hence, they 
may correspond to value function weights of tasks which are related to the input tasks. Hence, each node in the SOM could be assumed to correspond to a source task, and the SOM weight vector associated with an appropriately selected node could serve as source value function weights which could be used to guide the exploration of the agent while learning a new task. The details of the transfer mechanism are discussed in Section~\ref{TLalgo}.

In a continual learning scenario, 
an agent may encounter a number of tasks as it interacts with its environment. As per the metric defined in Eq.~\eqref{eqn1}, the value function weights corresponding to some of these tasks may possess a large degree of similarity, while others may vastly differ from each other. Generally, a SOM would be able to extract representative features in the value function weights of highly similar tasks. Learning and storing these representative features could help avoid the storage of redundant task knowledge.
However, a SOM containing only a few number of nodes may not be able to represent a wide range of task knowledge to a sufficient level of accuracy. Hence, the size of the SOM may need to adapt dynamically as and when new tasks are learned, and existing task knowledge is updated.
We address this problem by allowing the number of nodes in the SOM to change, using a mechanism similar to that used in the GSOM algorithm. For a SOM containing $N$ nodes, each node $i$ is associated with an error $e_i$ such that for a particular input vector $\vec{w}_{v_{j}}$, if node $s_{*}$ (with a corresponding weight vector $\vec{w}_{s_{*}}$) is the winner, the error $e_{s_{*}}$ is updated as:
\begin{equation}
\label{eqn2}
e_{s_{*}}\leftarrow e_{s_{*}}+1-c_{w_{v_{j}},w_{s_{*}}}.
\end{equation}
The term $(1-c_{w_{v_{j}},w_{s_{*}}})$ in Eq.~\eqref{eqn2} is proportional to the Euclidean distance between the $L^{2}$-norm versions of input vectors $\vec{w}_{v_{j}}$ and $\vec{w}_{s_{*}}$. Hence, the error update equation (Eq.~\eqref{eqn2}) is equivalent to that used in Alahakoon et al. \citep{alahakoon2000dynamic}. Once all the input vectors are presented to the SOM, the total error, $E$ of the network is simply computed as $E=\sum\limits_{i=1}^N e_{i}$.
The total error is computed for each iteration of the SOM. In subsequent iterations, if the increase in the total error per node exceeds a certain threshold $G_{T}$, new nodes are spawned at the boundaries of the SOM. Hence, growth of the SOM takes place if:

\begin{equation}
\label{somgrowtheqn}
\frac{\sum\limits_{i=1}^{N'} {e_{i}}^{k+1} -\sum\limits_{i=1}^N {e_{i}}^{k} }{N'}>G_{T},
\end{equation}\\
where ${e_{i}}^{k}$ is the error corresponding to node $i$ in iteration $k$, and $N'$ (where $N'\geq N$) is the number of nodes in the SOM in the subsequent iteration $k+1$. 

In our implementation, the configuration of the SOM is restricted to be square, and SOM growth occurs by adding new nodes only to the eastern (right) and southern (bottom) sides of the SOM. The weight vectors
of the newly spawned nodes are initialized to the mean of their neighbors, and are subsequently modified by the SOM training process. The tendency of this SOM training is to reduce the overall network error by achieving more accurate representations of the inputs presented to it. If the value functions are poorly represented, the average network error grows, until it exceeds the threshold $G_{T}$, which results in the growth of the SOM, as per Eq.~\eqref{somgrowtheqn}. In this way, the SOM can grow in size and representation capacity, 
while avoiding the storage of redundant task information. The avoidance of redundancy is supported by the fact that when the value functions of tasks that are highly similar to the SOM nodes are presented to the SOM, it does not spawn new nodes in response to this. New nodes are only spawned when the network fails to sufficiently represent the value function of the previously learned tasks. The overall GSOM training process is described in Algorithm \ref{alg:algorithm1}.

The nature of the described SOM algorithm is such that all the input vectors are needed during the training. However, for applications such as robotics, where the agent may have limited on-board memory, this may not be a feasible approach. Thousands of tasks may be encountered during its lifetime, and the value function weights of all these tasks would need to be explicitly stored in order to train the SOM. 
Ideally, we would like the knowledge contained in the SOM to adapt in an online manner, to include relevant information from new tasks as and when they are learned. We achieve this online adaptation by making modifications to the manner in which the SOM algorithm is trained. Specifically, when a new task is learned, we update the SOM by presenting the newly learned weights, together with the weight vectors associated with the nodes of the previously learned SOM as inputs to the GSOM algorithm. The resulting SOM is then used for transfer. In summary, the weights of the SOM are recycled as inputs while updating the knowledge base using the GSOM algorithm. The implicit assumption is that the weight vectors learned by the SOM sufficiently represent the knowledge of the previously learned tasks. This approach of updating the SOM knowledge base allows new knowledge to be adaptively incorporated into the SOM, while obviating the need to explicitly store the value function weights of all previously learned tasks.

\begin{algorithm}[tph]
  \caption{GSOM training mechanism}
  \begin{algorithmic}[1]
    \STATE \textbf{Inputs}: \STATEx $\mathbf{w_{v}}=\{\vec{w}_{v_{1}},..\vec{w}_{v_{i}},..\vec{w}_{v_{M}}\}:$~Input vectors to the GSOM algorithm. These may be value function weights of previously learned tasks or weights corresponding to the nodes of a previously learned SOM. \STATEx $N:$~Initial number of nodes in the SOM
\STATEx $\sigma_{0}:$ Initial value of neighborhood function $\sigma$
\STATEx $\tau_{1}:$ Time constant to control the neighborhood function
\STATEx $\kappa_{0}:$ Initial value of SOM learning rate $\kappa$
\STATEx $\tau_{2}:$ Time constant to control the learning rate
\STATEx $\mathbf{w_{s}}=\{\vec{w}_{s_{1}},..\vec{w}_{s_{i}},..\vec{w}_{s_{N}}\}:$~Initial weight vectors associated with the $N$ nodes in the SOM
    \STATEx $e:$ Error vector, initialized to be zero vector of length $N$
        \STATEx $E=0:$ Initial  value of average error
                \STATEx $G_{T}:$ Growth threshold parameter
                \STATEx $N_\text{iter}:$ Number of SOM iterations             
    \FOR {$i=1:N_\text{iter}$}
    \STATE Randomly pick an input vector $\vec{x}$ from $\mathbf{w_{v}}$ 
    \STATE Select winning node $n_{\mathrm{win}}$ based on highest cosine similarity to input vector $\vec{x}$
    \STATE $\sigma=\sigma_{0}\exp(-i/\tau_{1})$
    \STATE $\kappa=\kappa_{0}\exp(-i/\tau_{2})$
    \FOR {$j=1:N$}
    \STATE Compute topological distance $d_{n_{\mathrm{win}},j}$ between nodes $n_{\mathrm{win}}$ and $j$ 
    \STATE $h(n_{\mathrm{win}},j)=\exp(-d_{n_{\mathrm{win}},j}/2\sigma^{2})$
    \STATE $\vec{w}_{s_{j}}=\vec{w}_{s_{j}}+\kappa *h(n_{\mathrm{win}},j)*\|\vec{x}-\vec{w}_{s_{j}}\|$
    \ENDFOR
    \STATE $e(n_{\mathrm{win}})=e(n_{\mathrm{win}})+1-c_{x,w_{s_{n_{\mathrm{win}}}}}$
    \STATE $E_{i}=\sum_{k=1}^{N} e_{k}$
    \IF {$(E_{i}-E_{i-1})/N>G_{T}$}
    \STATE Trigger SOM growth: Spawn new SOM nodes and expand the error vector, with the values of new elements initialized to the mean of the previous error vector.
    \STATE Update $N$ as per the number of new nodes added
    \ENDIF
        \ENDFOR
  \end{algorithmic}
  \label{alg:algorithm1}
\end{algorithm}



\subsubsection{SOM Growth}
In Algorithm \ref{alg:algorithm1}, the nature in which the growth of the SOM occurs is not specified. Ideally, the growth must take place such that the SOM accurately summarizes the learned task knowledge, while also generalizing to tasks that are similar in nature. The growth should be measured in nature, only occurring when the current SOM is not able to appropriately represent the learned task knowledge. 
For the case where growth has just occurred ($N'>N$), if we assume the errors corresponding to the $N$ original nodes to be approximately the same across subsequent iterations of the GSOM training, then Eq.~\eqref{somgrowtheqn} can be written as:
\begin{align*}
\frac{\sum\limits_{i=1}^{N} {e_{i}}^{k+1} +\sum\limits_{i=N+1}^{N'} {e_{i}}^{k+1} -\sum\limits_{i=1}^N {e_{i}}^{k} }{N'}\leq G_{T}
\end{align*}\\
and hence,
\begin{align*}
\frac{\sum\limits_{i=N+1}^{N'} {e_{i}}^{k+1} }{N'}\leq G_{T}.
\end{align*}
If $e_{a}$ represents the average error associated with a node, then:
\begin{equation}
\label{avgerr_eq}
\frac{e_{a}(N'-N)}{N'}\leq G_{T} 
\Rightarrow e_{a}\leq \frac{N'G_{T}}{N'-N}.
\end{equation}\\
The maximum permissible average error $e_{a_\text{max}}$ for which further growth does not occur is thus: 
\begin{align*}
e_{a_\text{max}}=\frac{N'G_{T}}{N'-N}.
\end{align*}\\
The rate of change of this permissible quantity with respect to the size of the SOM network can then be derived to be:
\begin{equation}
\label{debydN_eqn}
\frac{d}{dN}(e_{a_\text{max}})=G_{T}\frac{N'-N\frac{dN'}{dN}}{(N'-N)^2}.
\end{equation}\\
The stationary point obtained by setting the right hand side of Eq.~\eqref{debydN_eqn} to zero gives us the update rule: $N'=KN$, where $K$ is a constant. In this case, since the number of SOM nodes must be an integer, $K$ is an integer. This solution, however, is neither a maximum nor a minimum, as  $\frac{d^2}{dN^2}(e_{a_\text{max}})=0$. However, it is interesting, as setting $N'=KN$ in Eq.\eqref{avgerr_eq} results in $e_{a}$ becoming dependent only on $G_{T}$ and $K$, and independent of $N$, the size of the SOM. Hence, this solution corresponds to the case where the maximum permissible value for $e_{a}$ is constant, and depends on $K$, and it can be shown that $\lim_{K\to\infty} e_{a_\text{max}}=G_{T}$. This is a useful property, as it  imposes a finite bound on $e_{a}$, and further SOM growth occurs only if $e_{a}$ exceeds this bound. 
However, the growth update rule $N'=KN$ falls short in terms of the convenience of implementation, as it does not specify the topology of the SOM. Specifically, the $KN$ nodes obtained after the SOM growth could be configured in a number of rectangular and non-rectangular topologies.

\begin{figure}[h]
\centering
\includegraphics[width=1\linewidth]{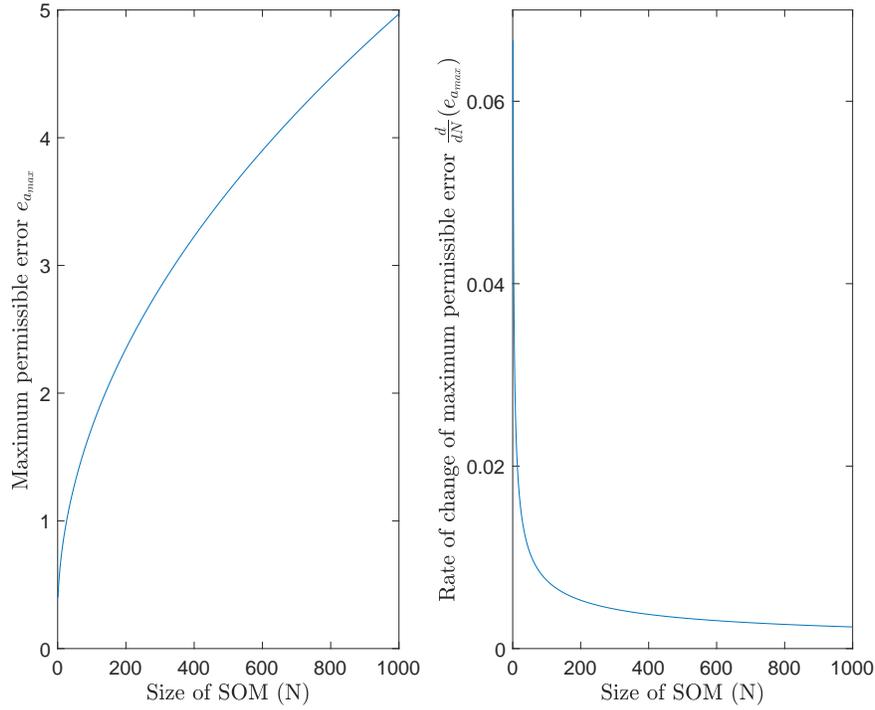}
\caption{Variations of $e_{a_\text{max}}$ and $\frac{d}{dN}(e_{a_\text{max}})$ with the size $N$ of the SOM.
} 
\label{limits}
\end{figure}

A convenient solution is to restrict the SOM to be square, such that the growth update rule is set to be $N'=(\sqrt{N}+1)^2$. By substituting this relation in Eq.~\eqref{avgerr_eq} and \eqref{debydN_eqn}, we obtain:
\begin{align*}
e_{a}\leq G_{T}\frac{(\sqrt{N}+1)^2}{1+2\sqrt{N}},
\end{align*}\\
and
\begin{align*}
\frac{d}{dN}(e_{a_\text{max}})=G_{T} \frac{1+\sqrt{N}}{(1+2\sqrt{N})^2}.
\end{align*}\\
Using these relations, the variations of $e_{a_\text{max}}$ and $\frac{d}{dN}(e_{a_\text{max}})$
 can be examined for the case when the SOM is always square (i.e., using the update rule $N'=(\sqrt{N}+1)^2$). Specifically, it is observed that $e_{a_\text{max}}$ and $\frac{d}{dN}(e_{a_\text{max}})$ respectively grows and diminishes as $O(\sqrt{N})$. Additionally, their asymptotic limits as $N\to\infty$ can be shown to be: $$\lim_{N\to\infty} e_{a_\text{max}}=\infty,$$ and $$\lim_{N\to\infty}\frac{d}{dN}(e_{a_\text{max}})=0.$$ 
 These trends are depicted in the Fig. \ref{limits}, which shows that the maximum permissible limit for the average error $e_{a}$ increases with the number of nodes, and the rate of increase decreases, and becomes nearly constant for larger values of $N$. Larger permissible limits of $e_a$ make it less likely for the SOM to grow further. However, large errors also imply the presence of SOM nodes which do not accurately represent its inputs.
While a less accurate SOM is undesirable, it also allows for greater diversity in the stored knowledge, which could potentially be beneficial for guiding the learning of target tasks when they are highly dissimilar to the previously learned tasks. Moreover, as previously mentioned, restricting the topology to be square is superior with respect to preventing runaway growth of the SOM, making it a scalable approach for knowledge storage. 

\subsection{Transfer Mechanism}
\label{TLalgo}
Once the knowledge of previously learned tasks has been assimilated into a SOM, it is reused to aid the learning of a target task.  The weight vector associated with each node in the SOM is treated as the value function weight vector corresponding to an arbitrary source task. 
Among these source value function weight vectors ($\mathbf{w_{s}}$), the one that is most similar to the target value function weight vector $w_{T}$ is chosen for transfer. That is, the index of the most similar source task is given by:
$$s_{*}=\underset{i \in \mathbb{N}^{N}_{>0}}{\operatorname{argmax}c_{{w_{i}},{w_{T}}}},$$
and the corresponding source value function weight vector used for transfer is $w_{s_{*}}$. Here, $\mathbb{N}^{N}_{>0}$ is the set of all positive natural numbers up to $N$. 

It must be noted that the relevance of the selected weight vector $w_{s_{*}}$ for transfer depends on how well $w_{T}$ has been estimated. For example, compared to a randomly initialized $w_{T}$, a partially converged $w_{T}$ would be more likely to pick out an appropriate source weight vector from $\mathbf{w_{s}}$, such that it is capable of providing action advice relevant to the target task being learned.

In addition to biasing the exploratory actions, transfer could also possibly be achieved by allowing the selected source task weights to directly modify the value function weights of the target task. 
This could be done, for instance, by biasing the target value function weights to be closer to the selected source task weights.
However, for a particular task, some of the elements of the weight vector may have a greater influence on the agent's behavior in comparison to others. The cosine similarity measure does not capture such asymmetries in the sensitivities of the weight vector elements. 
Hence, the direct influence of the selected source task weights on the weight parameters of the target task could be detrimental to the agent's target task performance. 
In contrast to this, our approach of allowing the selected source value function weights to guide the exploratory actions of the agent is a subtler, and hence, safer approach for biasing the value function of the target task.

\begin{algorithm}[h]
  \caption{The Transfer Mechanism}
  \begin{algorithmic}[1]
  \STATE \textbf{Inputs}:
  \STATEx trained $SOM$ with $N$ nodes, corresponding to $N$ source value function weights $\mathbf{w_{s}}=\{\vec{w}_{s_{1}},..\vec{w}_{s_{i}},..\vec{w}_{s_{N}}\}$
  \STATEx Target task $T$, initialized with a value function weight $w_{T}$
  \STATEx $N_E$: Maximum number of $Q$-learning episodes
  \FOR{$i=1:N_{E}$}
  \WHILE{terminal state is not reached}
  \STATE $s_{*}=\underset{i \in \mathbb{N}^{N}_{>0}}{\operatorname{argmax}c_{{w_{i}},{w_{T}}}}$, where $s_{*}$ is the index of the winning node
    \STATE With probability of $1-\epsilon$, choose action $a$ to be greedy with respect to $w_{T}$, and with a probability of $\epsilon$, let $a$ be greedy with respect to $w_{s_{*}}$.
    \STATE Update $w_{T}$ using standard $Q$-learning update equation.
  \ENDWHILE
  \ENDFOR
  \STATE Update SOM as per Algorithm \ref{alg:algorithm1}, using $w_{T}$ as one of the input vectors 
  \end{algorithmic}
  \label{alg:transfermechanism}
\end{algorithm}


\subsection{Adaptive Clustering for Multi-task Learning}
\label{adapt_clust}

In the navigation experiments described in Section~\ref{results}, in order to provide agents with a greater degree of autonomy with respect to choosing their goals, we allow goal locations in the environment to be automatically discovered by the agent itself. This is achieved by simply applying an approach described in Karimpanal et al. \citep{KARIMPANAL201739}, where an environment feature vector $\vec{F_{e}}$ is defined, and unique configurations of this feature vector are discovered using an adaptive clustering algorithm. These discovered clusters are treated as the feature vectors associated with the goal locations of arbitrary tasks, which are then learned in parallel (that is, multiple value function weights are updated with each interaction) using off-policy learning algorithms such as $Q$-learning.

As the agent moves through the environment, it senses feature vectors $\vec{F_{e}}$, and the clustering algorithm assigns them to different clusters, based on their Euclidean distances with the centroids of the different clusters. 
Next, the element-wise absolute distance between the centroid of the assigned cluster and components of $\vec{F_{e}}$ is computed. For each element, if this distance lies within a certain number of standard deviations of the corresponding element in the centroid, then $\vec{F_{e}}$ is considered to belong to that cluster; if not, a new cluster is seeded. Each new cluster is seeded with an initial non-zero variance, in order to maintain a certain level of uncertainty about the cluster centroids. The uncertainty reduces as more numbers of  samples are observed. Each time a cluster receives a new member, the centroid and variance of each of the $j^{th}$ feature element in the cluster is updated online using the corresponding elements of  $\vec{F_{e}}$, as follows:

\begin{equation*}\label{Ch1:meanupdate}
\nu_{j}\longleftarrow(N_{C}*{\nu}_{j}+F_{e}^{j})/(N_{C}+1)
\end{equation*}

\begin{equation*}\label{Ch1:varupdate}
{Var_{j}}\longleftarrow (N_{C}*(Var_{j} +\nu_{j}^2)+{F_{e}^{j}}^2)/(N_{C}+1)-\nu_{j}^2
\end{equation*}

\begin{equation*}\label{Ch1:membersupdate}
N_{C}\longleftarrow N_{C}+1
\end{equation*}
where $\nu_{j}$ and ${Var_{j}}$ are respectively the mean (centroid) and variance of the $j^{th}$ feature element in the cluster, and $N_{C}$ is the number of members in cluster $C$. In this way, the approach serves to cluster the feature space in an unsupervised and adaptive manner without prior knowledge of the number of clusters that exist in the space. Each cluster centroid is then treated as the environment feature vector associated with an arbitrary task in the environment. Doing so enables these tasks to be learned simultaneously using off-policy algorithms.

The purpose of allowing agents to learn multiple tasks in this off-policy manner is so that they are equipped with some priors for the value functions of the different tasks in its environment. Such a prior, if acquired for a particular task, could provide a basis for the initial selection of source tasks from the SOM, when the value function of the corresponding task is being learned. In addition, this approach of autonomously discovering and learning tasks equips the agents in Section~\ref{results} with more autonomy and better life-long learning \citep{ring1994continual} abilities. The SOM based knowledge storage and transfer approaches described in Sections~\ref{SOM_approach} and~\ref{TLalgo}, are however, independent of this autonomous task identification approach, and are intended to be applicable in a more general sense. 

\section{Results}
\label{results}
We use the knowledge storage and reuse mechanisms described in Section \ref{method} to accelerate the learning of target tasks in navigation environments. We implement the described mechanisms in simulation as well as with actual experiments using a micro-robotics platform.
The details of these implementations are described in this section.

\subsection{Simulation Experiments}
\label{simulations}
\begin{figure}[h]
\centering
\includegraphics[width=0.8\linewidth]{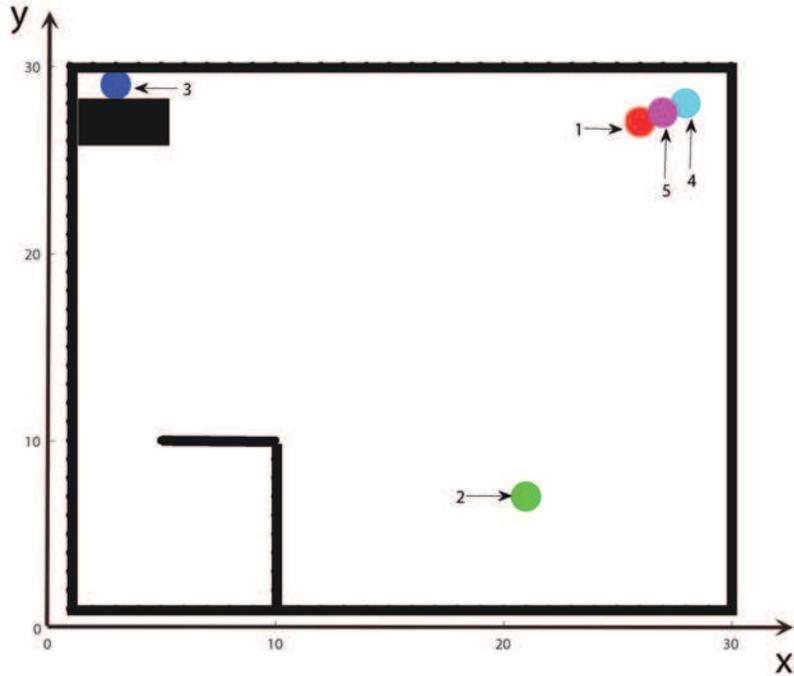}
\caption{The simulated continuous environment with the navigation goal states of different tasks (numbered from tasks $1$ to $5$), indicated by the different colored circles.   
} 
\label{env}
\end{figure}

In order to evaluate the described knowledge storage and reuse mechanisms, we allow the agent to explore and learn multiple tasks in the simulated environment shown in Fig.~\ref{env}. The environment is continuous, and the agent is assumed to be able to sense its $x$ and $y$ coordinates, which constitute its state. The states are represented in the form of a binary feature vector $\vec{F_{a}}$ containing $100$ elements for each state dimension. While navigating through the environment, the agent is allowed to choose from a set of $9$ different actions: moving forwards, backwards, sideways, diagonally upwards or downwards to either side, or staying in place. The speeds associated with these movements is set to be 6 spacial units/s, and new actions are executed every 200 ms.

As the agent executes actions in its environment, it autonomously identifies tasks using the adaptive clustering approach described in Section~\ref{adapt_clust}. The clustering is performed on the environment feature vector $\vec{F_{e}}$, which contains elements describing the presence or absence of specific environment features. For instance, these features could represent the presence or absence of a source of light, sound or other signals from the environment that the agent is capable of sensing.
In the simulations described here, the environment feature vector $\vec{F_{e}}$ contains $4$ elements corresponding to $4$ arbitrary environment stimuli distributed at different locations in the environment. As the agent interacts with its environment, clustering is performed on $\vec{F_{e}}$ in an adaptive manner, which helps identify unique configurations of $\vec{F_{e}}$ which may be of interest to the agent. During the agent's interactions with the environment, the mean of each discovered cluster is treated as the environment feature vector associated with the goal state of a distinct navigation task.  In our simulations, the agent eventually discovers $5$ such tasks, the corresponding goal locations of which are indicated by the colored regions in Fig.~\ref{env}. The value function corresponding to each of these tasks is learned using $Q$-learning with linear function approximation \citep{sutton2011reinforcement}. For $Q$-learning, the reward structure is such that the agent obtains a reward ($+100$) when it is in the goal state, a penalty ($-100$) for bumping into an obstacle, and a living penalty ($-10$) for every other non-goal state. In each episode, the agent starts from a random state and executes actions till it reaches the associated navigation target region (goal state), at which point, a positive reward is obtained, and the episode terminates. For each $Q$-learning task, the full feature vector $\vec{F}$ (where $\vec{F}=\{\vec{F_{e}}\cup \vec{F_{a}}\}$) is used, and the learning rate $\alpha$ is set to be $0.3$, the discount factor $\gamma$ is $0.9$ and the trace decay parameter $\lambda$ is set to be $0.9$. The other hyperparameters described in Algorithm \ref{alg:algorithm1} are set to the following values for both the simulations and experiments in this work: $N=4$, $\sigma_{0}=50$, $\tau_{1}=250$, $\tau_{2}=0.1$, $G_{T}=0.3$ and $N_\text{iter}=1000$.

Once a new navigation task $T$ is identified, and its value function weight vector $w_{T}$ is learned, we incorporate this new knowledge into the SOM knowledge base. In order to do this, the value function weight vector associated with the newly learned task, along with the weight vectors associated with the SOM are presented as input vectors to Algorithm~\ref{alg:algorithm1}. For instance, if the weight vectors of the SOM are given by $\mathbf{w_{s}}=\{\vec{w}_{s_{1}},..\vec{w}_{s_{i}},..\vec{w}_{s_{N}}\}$, then the subsequent input vectors $\mathbf{w_{\mathrm{v}}}$ to Algorithm~\ref{alg:algorithm1} are $\mathbf{w_{\mathrm{v}}}=\{\mathbf{w_{s}}\cup \vec{w}_{T}\}$. By presenting the inputs to the GSOM algorithm in this manner, the resulting SOM approximates and integrates previously learned task knowledge and the knowledge of newly learned tasks. 

\begin{figure}[h]
\centering
\begin{subfigure}{.5\textwidth}
  \centering
  \includegraphics[width=1\linewidth]{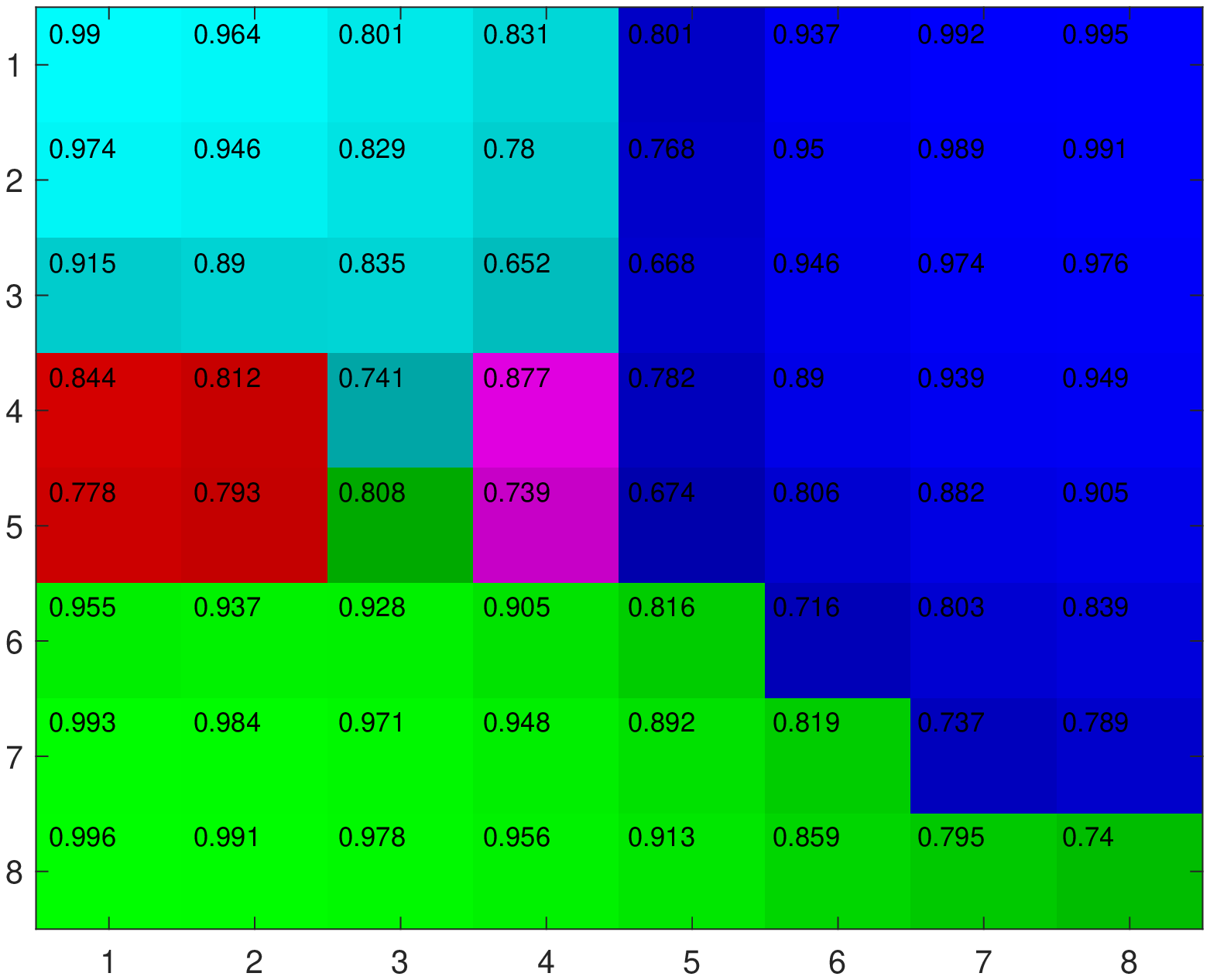}
  \caption{}
  \label{fig:8by8}
\end{subfigure}%
\begin{subfigure}{.5\textwidth}
  \centering
  \includegraphics[width=1\linewidth]{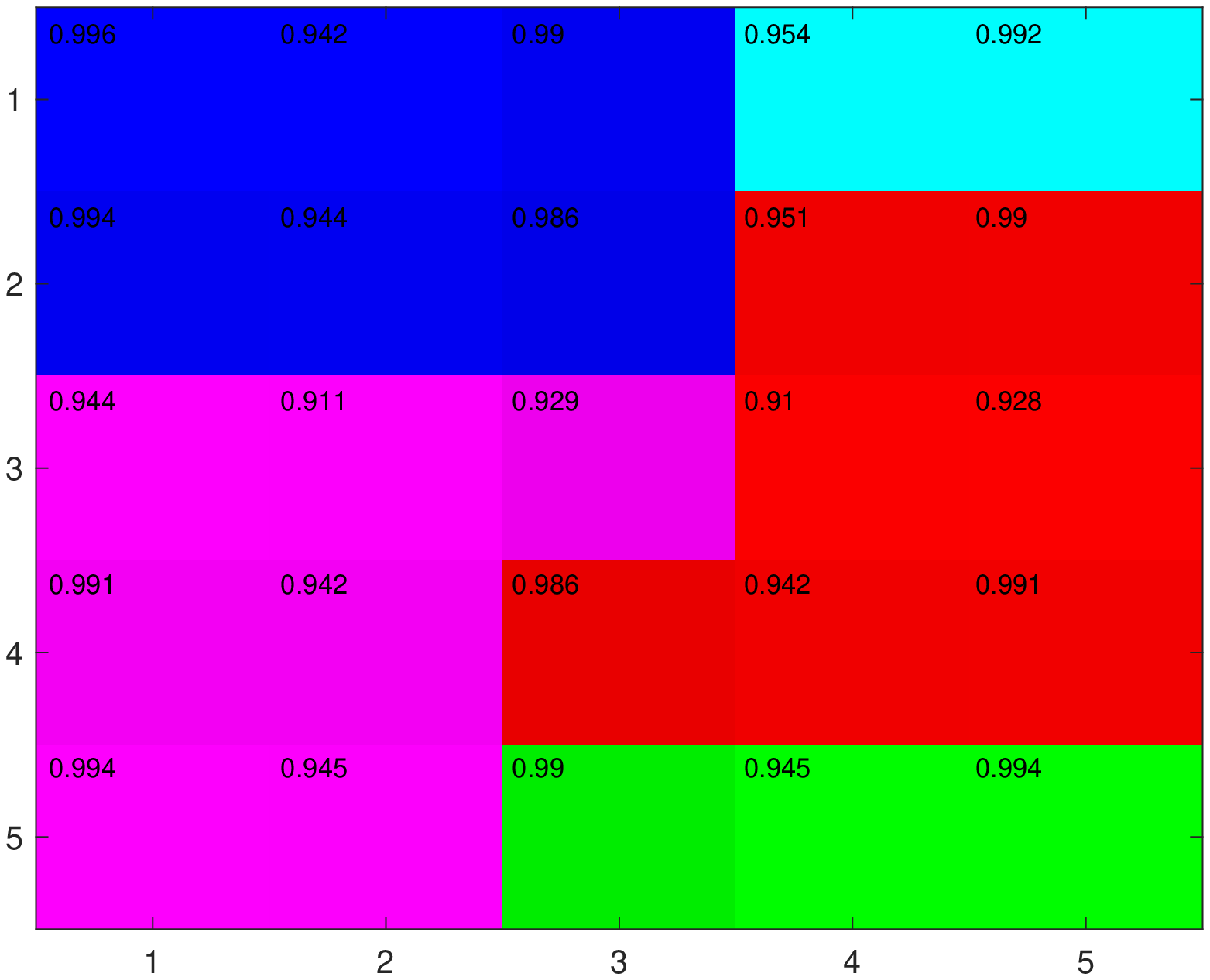}
  \caption{}
  \label{fig:5by5}
\end{subfigure}
\caption{(a) A visual depiction of an $8\times 8$ SOM resulting from the simulations in Section \ref{simulations}, where value functions are represented using linear function approximation. (b) Shows a $5 \times 5$ SOM which resulted when the simulations were carried out using a tabular approach. In both (a) and (b), the color of each node is derived from the most similar task in Fig.~\ref{env}. The intensity of the color is in proportion to the value of this similarity metric (indicated over each SOM element).}
\label{fig:SOMmap}
\end{figure}

\begin{figure}[h]
\centering
\includegraphics[width=0.8\linewidth]{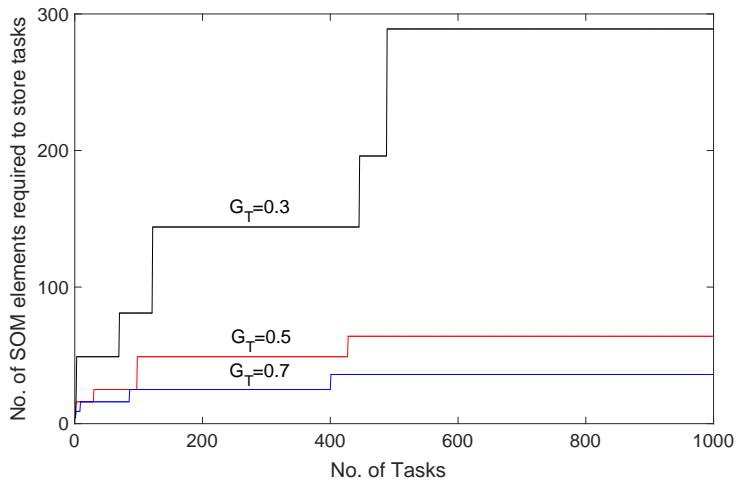}
\caption{The number of SOM nodes used to store knowledge for up to $1000$ tasks, for different values of growth threshold $G_{T}$.
} 
\label{fig:SOMscaling}
\end{figure}

Fig.~\ref{fig:8by8} shows a sample $8\times 8$ SOM, which was learned by the agent after $1000$  $Q$-learning episodes. Similarly, Fig.~\ref{fig:5by5} shows a $5\times 5$ SOM which resulted from a tabular approach to the same navigation problem. This demonstrates the flexibility of this approach with respect to different representation schemes. Although these SOMs store more value functions than the number of tasks, as demonstrated later on (using  Fig.\ref{fig:SOMscaling}), the representation becomes more storage efficient when a large number of tasks are involved.
The color of each SOM element in Fig.~\ref{fig:SOMmap} corresponds to the task in Fig.~\ref{env} that has the maximum cosine similarity between its value function weights and the weight vector associated with that SOM element. Further, the brightness of this color is in proportion to the value of this cosine similarity. In Fig.~\ref{fig:SOMmap}, these values are overlaid and displayed on top of each SOM element. The distribution of the different colors and associated cosine similarity values of each SOM element in Fig.~\ref{fig:SOMmap} suggests that the SOM stores knowledge of a variety of related tasks.
Specifically, Fig.~\ref{fig:SOMmap} shows that the nodes corresponding to tasks that have very different goal locations (measured perhaps by how far apart they are in physical space) form separate, distinct clusters (for example, the blue and green clusters in the SOM, representing nodes related to tasks $2$ and $3$). In contrast, nodes corresponding to tasks whose goal locations are close to each other (such as tasks $1$, $4$ and $5$) are generally never too far away from each other in the map (as inferred from the locations of the red, cyan and pink clusters). 
This shows that the allocation of the SOM nodes is done as per the characteristics of the tasks, and not merely according to the number of tasks. The latter approach would result in significant redundancies, for example, if the agent encounters multiple tasks which are very similar to each other, or the same task multiple times. Such redundancies are avoided by the proposed SOM-based approach. 

\begin{figure}[h]
\centering
\includegraphics[width=0.8\linewidth]{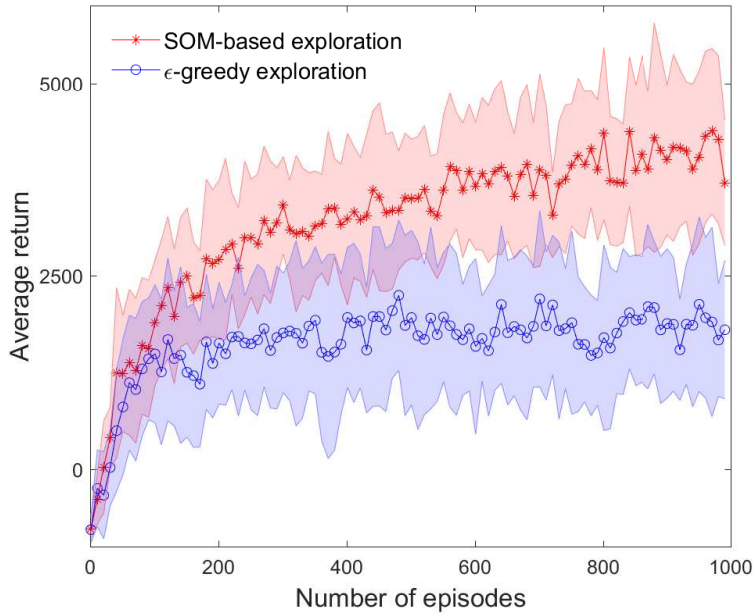}
\caption{A sample plot of the nature of the learning improvements brought about by SOM-based exploration (for $G_{T}=0.3$). The solid lines represent the mean of the average return for $10$ $Q$-learning runs of $1000$ episodes each, whereas the shaded region marks the standard deviation associated with this data.
} 
\label{fig:avgreturns}
\end{figure}

\begin{figure}[h]
\centering
\begin{subfigure}{0.5\textwidth}
  \centering
  \includegraphics[width=1\linewidth]{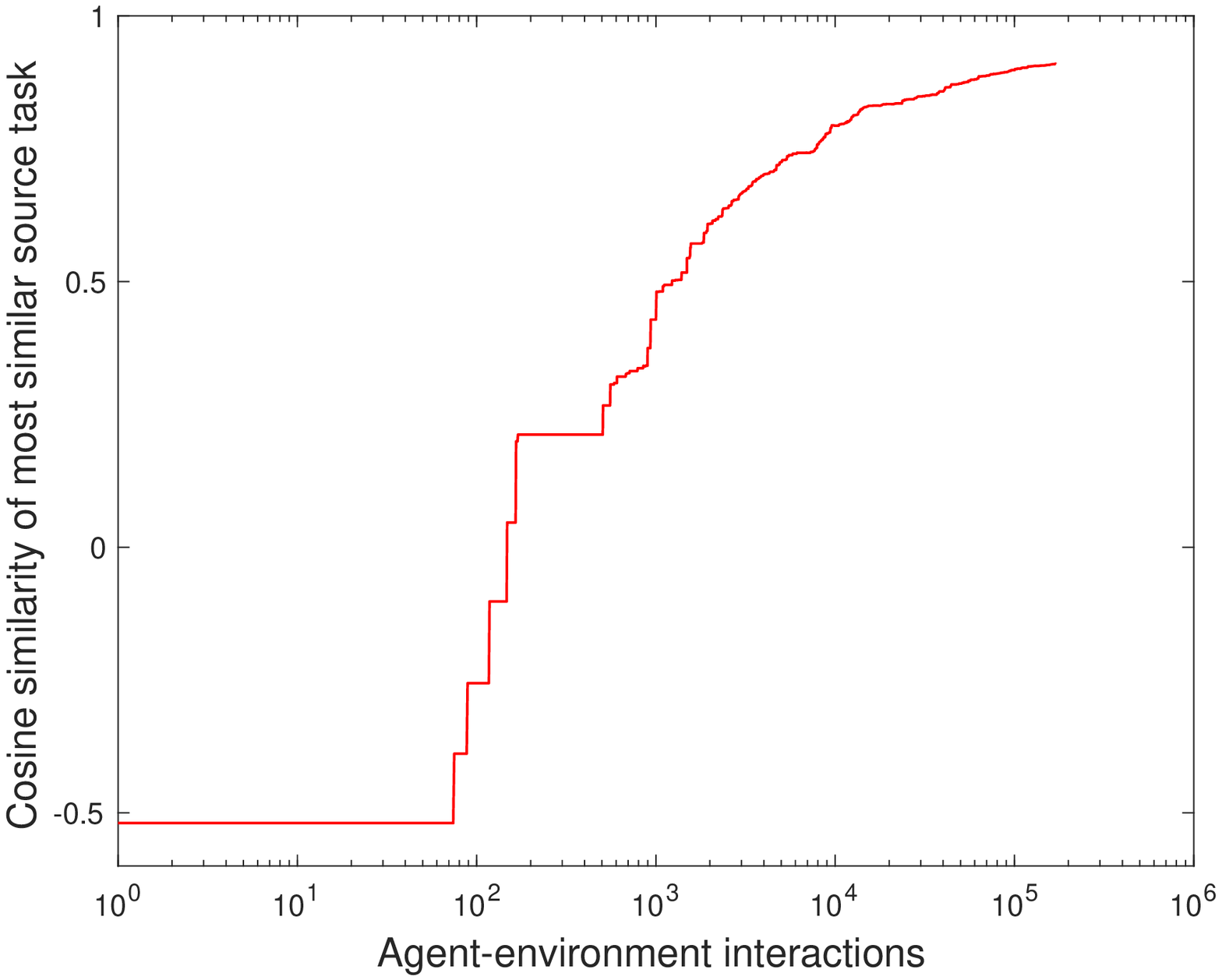}
  \caption{}
  \label{fig:SOMadvice}
\end{subfigure}%
\begin{subfigure}{.5\textwidth}
  \centering
  \includegraphics[width=1.05\linewidth]{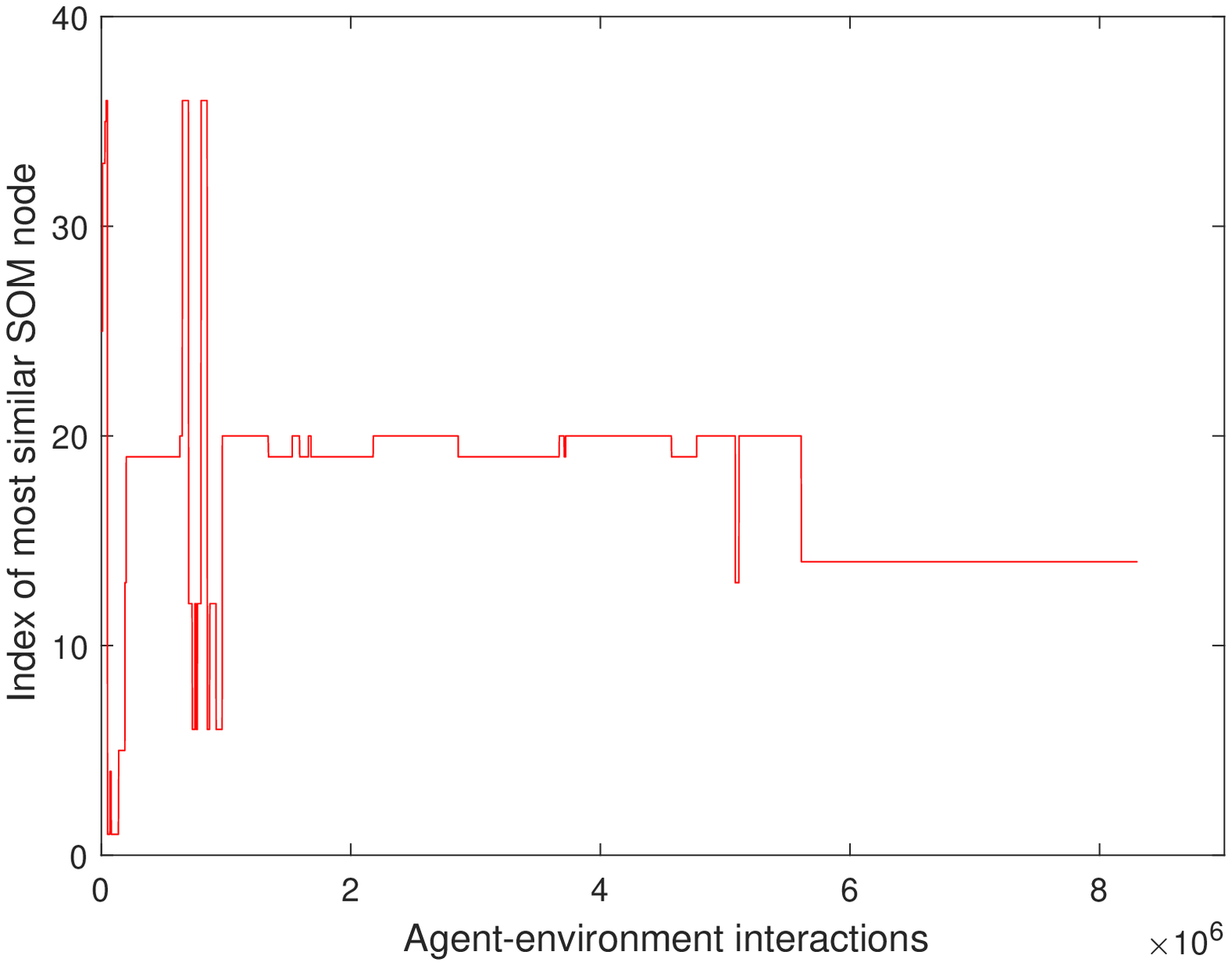}
  \caption{}
  \label{fig:win_indices}
\end{subfigure}
\caption{(a) A representative example of the variation of the cosine similarity between a target task and its most similar source task as the agent interacts with its environment. (b) An example of the variation of the index of the most similar SOM node as the agent interacts with the environment.}
\label{fig:SOM_interactions}
\end{figure}

Although the SOM knowledge base does not necessarily retain the exact value function weights of previously learned tasks, it can be used to efficiently guide the exploration of an agent while learning a new task. This is especially true if the new task is closely related to one of the previously learned tasks. Fig.~\ref{fig:avgreturns} depicts this phenomenon for task $5$ ($\epsilon=0.3$), with higher returns being achieved at a significantly faster rate using the SOM-based exploration strategy described in Section \ref{TLalgo}. In both exploration strategies (SOM-based and $\epsilon$-greedy), exploratory actions are executed with the same probability, but the SOM-based exploration achieves a better performance, as knowledge of related tasks (in this case, tasks $1$ and $4$) from previous experiences allows the agent to take more informed exploratory actions.  

This is also supported by the results in Fig.~\ref{fig:SOMadvice}, which shows the evolution of the cosine similarity between the value function weights of the target task and the most similar weight vector in the SOM as the agent interacts with its environment. With a greater number of agent-environment interactions, the estimates of the agent's target task weight vector improves, and it receives more relevant advice from the SOM. In addition to Fig.~\ref{fig:SOMadvice}, in Fig.~\ref{fig:win_indices}, we observe that the index of the most similar SOM node fluctuates significantly during the initial stages of learning, when the estimate of the target value function weights is poor. As vastly different indices generally correspond to different regions in the SOM (and hence value functions that are very different in nature), this implies that the initial exploratory advice provided by the SOM is mostly random. As the learning progresses, the target value function estimate improves and stabilizes, and the most similar SOM node consistently occurs around a particular topological neighborhood of the SOM map. This is revealed by the lack of drastic fluctuations in the latter portions of Fig.~\ref{fig:win_indices}. These trends suggest that the quality of advice derived from the SOM improves with the number of agent-environment interactions, which leads to the learning improvements seen in Fig.~\ref{fig:avgreturns}. 

As observed in Fig.~\ref{fig:avgreturns}, our approach does not lead to sudden, dramatic jumpstart improvements, as the transfer is solely based on using the SOM to take more informed exploratory actions. Although our approach may limit the bias that could potentially be added for learning a target task, it ensures against drastic drops in the learning performance. This is because each target task is learned from scratch, and  improvements are brought about only through improved exploratory actions, whose influence on the value functions is subtler in comparison to the approach of directly modifying the value function weight parameters.

\begin{figure}[h!]
\centering
\includegraphics[width=0.8\linewidth]{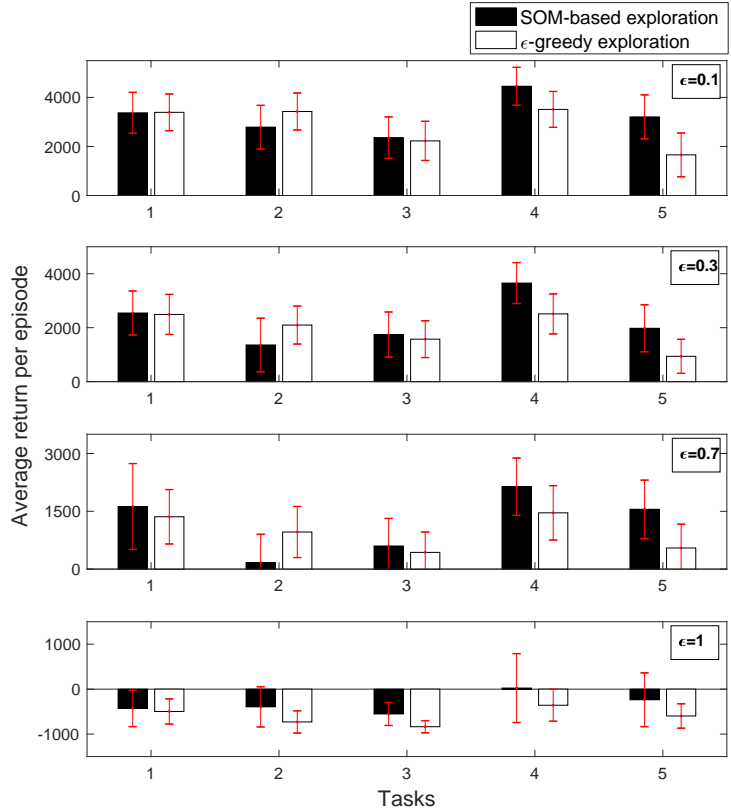}
\caption{Comparison of the average returns accumulated for different tasks in simulation using the SOM-based and $\epsilon-$greedy exploration strategies.} 
\label{fig:returnplot}
\end{figure}

Fig.~\ref{fig:returnplot} shows the average return per episode for different tasks and different values of $\epsilon$, using the two exploration strategies. The values plotted are averaged over $10$ runs. The return is computed through evaluation runs conducted after (as opposed to during) each episode by allowing the agent to greedily exploit the value function weights starting from $100$ randomly chosen points in the environment for $100$ steps. This allows us to examine the learning improvements even for highly exploratory strategies (for example, when $\epsilon=1$). As observed from Fig.~\ref{fig:returnplot}, SOM-based exploration consistently results in higher average returns for related tasks $4$ and $5$. Its performance on the unrelated tasks $2$ and $3$ are generally comparable to that of the $\epsilon-$greedy approach. Although task $1$ is related to tasks $4$ and $5$, it is the first task learned by the agent. So, it cannot make use of its previous knowledge to accelerate its learning on this task. Hence, the transfer advantage is not observed for task $1$. However, overall, it is useful to extract exploratory action advice from the SOM.

\begin{figure}[H]
\centering
\includegraphics[width=0.8\linewidth]{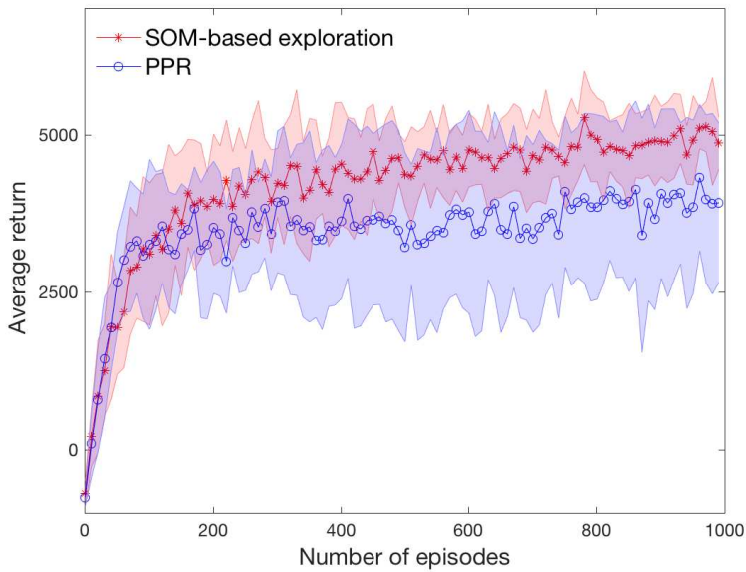}
\caption{A comparison between the learning improvements brought about by SOM-based exploration and the PPR approach for target task $5$. The solid lines represent the mean of the average return for $10$ $Q$-learning runs of $1000$ episodes each, whereas the shaded region marks the standard deviation associated with this data.
} 
\label{fig:prqplot}
\end{figure}

In order to put these described learning improvements into perspective, we also compared the transfer performance of our approach to that of the PPR algorithm, which was briefly mentioned in Section \ref{relatedwork}. To perform this comparison, we provided the agent with a set of policies (policies corresponding to tasks 1-4, which comprised a policy library) corresponding to learned navigation tasks in the environment described in Fig.~\ref{env}, and allowed it to learn a policy for task $5$. The new task was learned using the PPR algorithm, which made use of the policy library in order to guide its exploration. Subsequently, this task was independently learned again using our approach, by simply replacing the exploration strategy in the PPR approach with the proposed SOM-based exploration strategy. The SOM used for this was derived from the same set of policies in the mentioned policy library.
During these simulations, the PPR-related parameters were set as follows: initial exploration parameter $\psi=1$, decay rate of exploration parameter $\nu=0.95$, initial temperature parameter $\tau=0$ and step change in temperate parameter $\Delta\tau=0.05$, as specified in Fernandez et al.~\citep{fernandez2013learning}. The $Q$-learning parameters were left unchanged from the previous navigation tasks mentioned in this section. A comparison of the learning performance for the target task $5$, averaged over $10$ runs, is depicted in Figure~\ref{fig:prqplot}. As observed, the learning performance of the agent is superior when it employs the SOM-based exploration approach. This is probably due to the fact that unlike PPR, which solely exploits the past policies, the SOM-based approach exploits past policies as well as non-linear interpolations between these policies, which happen to correspond to policies that are useful for solving other tasks in the environment.


In addition to the learning improvements described, the described SOM-based transfer approach also offers advantages in terms of the scalability of knowledge storage. This is depicted in Fig.~\ref{fig:SOMscaling}, which shows the number of SOM nodes needed for storing the knowledge of up to $1000$ tasks, with different values of the GSOM threshold parameter $G_{T}$. It is clear that as the number of learned tasks increases, the number of SOM nodes required per task decreases, making the SOM-based approach more scalable with respect to knowledge storage. However, it should be noted that for a small number of tasks, the proposed SOM representation may not be efficient. Such an inefficiency is observed in Figure~\ref{fig:SOMmap}, where the number of nodes needed to store the knowledge of tasks is much larger than the number of tasks. Hence, the storage efficiency of the proposed approach becomes relevant, generally in cases where a large number of tasks are involved.

The simulation results in this section suggest that adopting the SOM-based exploration strategy may be beneficial for learning a new task which is related to previously learned tasks. Even when the new task is unrelated (such as in the case of tasks $2$ and $3$), employing such an exploration strategy does not lead to drastic reductions in performance. In Section \ref{exptsection}, we conduct knowledge storage and transfer experiments similar to those described in this section, in a real world navigation environment using a micro-robotics platform.

\subsection{Robot Experiments}
\label{exptsection}
\begin{figure}[h]
\centering
\includegraphics[width=1\linewidth]{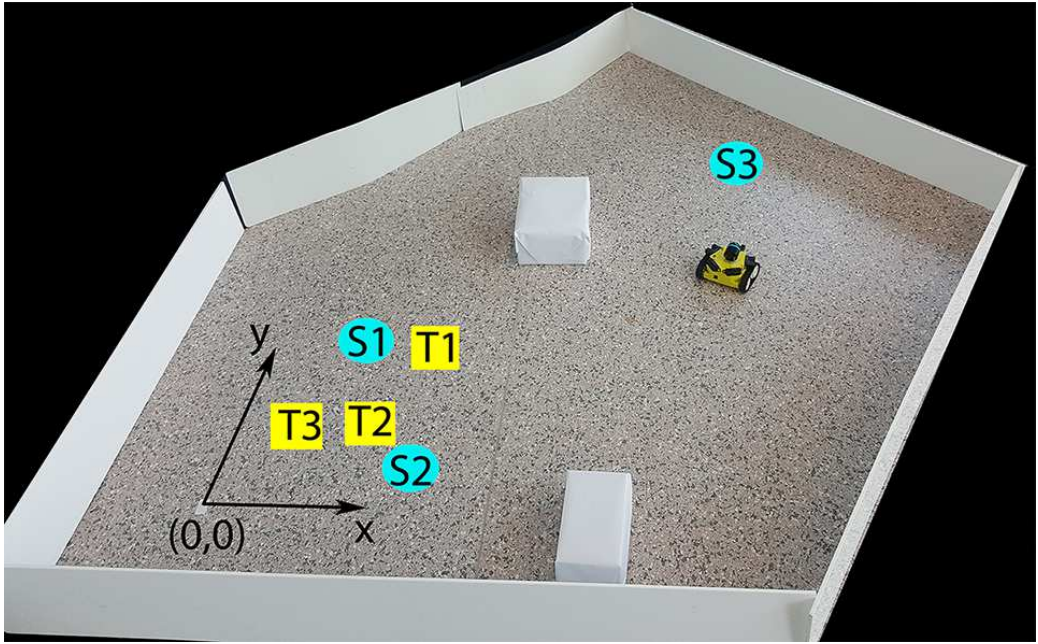}
\caption{The environment set-up and configuration, showing the position of the robot's coordinate axes, and the goal locations of the different identified tasks (S$1$, S$2$ and S$3$) and target tasks (T$1$, T$2$ and T$3$).} 
\label{fig:exptsetup}
\end{figure}

In this section, the methodology described in Section~\ref{method} is further validated with real world experiments using the EvoBot \citep{karimpanal2015adapting}, a mobile micro-robotics prototyping platform. The EvoBot is a differentially driven robot, and it uses wireless communication to exchange information with a central computer. The computer receives data from the robot's sensors, performs computations, and transmits a command for the robot to execute. The action set of the robot is composed of $5$ different actions: moving straight, curving left, curving right, spinning right and spinning left. To sense its surrounding environment, the robot is equipped with $3$ infrared sensors on its front side, each separated by an angular separation of $72\degree$ from the other. Apart from this, the robot also has a number of sensors for localization. An extended K\'alm\'an filter \citep{anderson1979optimal} combines these sensor readings to maintain a good estimate of the robot's position in its environment.

The experiments described in this section are carried out in an environment (approximately $1.8~\mathrm{m}\times1.8~\mathrm{m}$ in size) with coordinate axes fixed as shown in Fig.~\ref{fig:exptsetup}. The walls and obstacles in the environment are colored white in order for them to be more easily detected by the infrared sensors of the robot. The robot's state consists of its $x$ and $y$ coordinates, along with its orientation (heading direction) in the environment. Three locations in the environment (indicated by locations S$1$, S$2$ and S$3$ in Fig.~\ref{fig:exptsetup}) are assumed to be associated with the feature elements of the environment feature vector. For RL tasks in this environment, the feature vector is composed of $803$ feature elements ($300$ for each of the horizontal and vertical coordinates, $200$ for the heading, and the $3$ feature elements of the environment feature vector). As in Section \ref{simulations}, the environment feature vector is used for the identification of different tasks via clustering. 

For an RL task of navigating to a goal location in the environment shown, the reward structure is such that the robot receives a positive reward (arbitrarily set to $+100$) when it is within $10$~cm of the associated goal location and a living penalty ($-10$) for every non-goal state. Penalties of $-100$ are assigned to states in which the robot is too close to an obstacle. In order to avoid running into an obstacle, certain `safe' actions (actions which help steer the robot away from obstacles) are defined when any of the robot's infrared sensors detect an obstacle within $30$~cm of it. These actions are determined based on the infrared sensor readings of the robot. For instance, if the infrared sensor on the left of the robot reports an obstacle within $30$~cm, the safe actions could be curving or spinning right. In order to discourage unsafe actions, 
each time the robot comes close ($\leq 30$~cm) to an obstacle (where it receives a large penalty of $-100$), we ensure that non-safe actions do not result in any robot motion. Hence, when a non-safe action is selected, the robot remains in the undesirable state, and the value function is updated based on the large penalties it receives in that state. However, when safe actions are chosen, the robot is allowed to move out of the region associated with large penalties, and the reward it receives is relatively better than the penalty of $-100$. For both safe and unsafe actions, the value functions are updated as usual. The difference is that for unsafe actions, the reward is forced to be low by disallowing the robot's motion in the undesirable state. In this way, unsafe actions are discouraged, and over time, the robot becomes more likely to choose safe actions when it is close to an obstacle.

The robot is initially allowed to explore the environment for a period of $1$ hour with actions chosen at random (exploration parameter $\epsilon=1$) from the action set with a frequency of approximately $3$~Hz. During this exploration phase, the environment feature vectors are clustered in an adaptive manner, leading to the identification of different tasks (that is, tasks of navigating to points S$1$, S$2$ and S$3$). The knowledge of these identified tasks are used to construct the SOM knowledge base, which is later used to learn the target tasks (tasks corresponding to locations T$1$, T$2$ and T$3$, as shown in Fig.~\ref{fig:exptsetup}). The value function weights associated with each of these identified tasks are learned in parallel using $Q$-learning with linear function approximation. The parameters used for each $Q$-learning task are the same as those used in the simulations. A similar reward structure is used for all the $Q$-learning tasks, with the only difference being the locations associated with positive rewards.



Once the value function weights of the different identified tasks are learned, they are stored in a SOM using Algorithm \ref{alg:algorithm1}. The robot is then assigned to sequentially learn a series of target tasks using $Q$-learning with both the SOM-based and $\epsilon-$ greedy exploration strategies. These target tasks (T$1$, T$2$ and T$3$ tasks) are chosen such that their goal state is physically close to the goal states of at least some of the source tasks. The purpose of choosing target tasks in this manner is so that we may evaluate the 
learning performance of the robot for tasks that are related to those already learned by the robot. The hypothesis is that in the case of the SOM-based exploration, the robot will be able to leverage its knowledge of related tasks to appropriately guide its exploratory actions, leading to the accumulation of larger returns, compared to the case where exploratory actions are chosen at random. For each target task, the performance of the different exploration strategies (with $\epsilon=0.7$) is evaluated as the average sum of rewards (return) accumulated over $10$ runs, each of which lasts for a duration of $300$~s. 

\begin{figure}[h]
\centering
\includegraphics[width=0.8\linewidth]{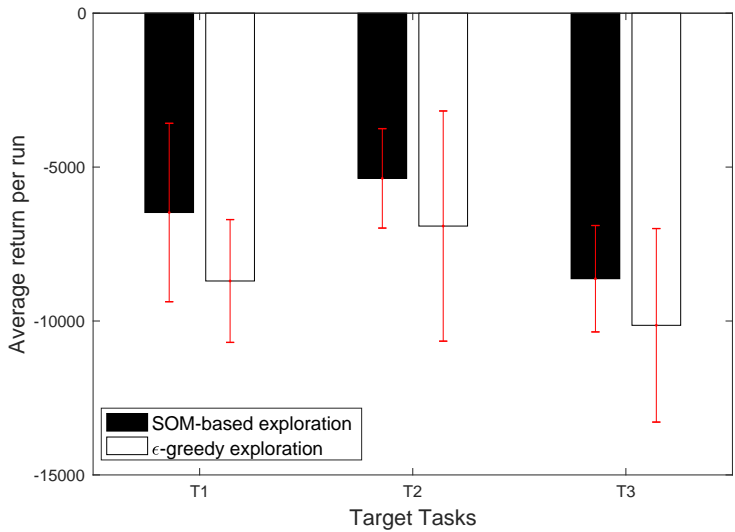}
\caption{Comparison of the average returns accumulated using SOM-based exploration and $\epsilon-$greedy exploration while learning the target tasks $\mathrm{T}1$, $\mathrm{T}2$ and $\mathrm{T}3$.} 
\label{fig:exptres}
\end{figure}

Fig.~\ref{fig:exptres} summarizes the comparison between the two exploration strategies. Given the relatively short time of $300$~s, the goal state need not be visited during every run. In addition to this, the environment is set up such that negative rewards are much more commonly experienced than positive ones. Owing to these factors, the sum of rewards (return) in all the runs is negative. However, SOM-based exploration is found to accumulate a higher average return as compared to the $\epsilon-$greedy exploration strategy. 
As the robot interacts with its environment, the estimates of its value function weights improve. When the SOM-based exploration strategy is employed, these improved estimates allow it to receive more relevant suggestions for exploratory actions (using the mechanism described in Section~\ref{TLalgo}) from the SOM knowledge base. This accounts for the improved performance observed in Fig.~\ref{fig:exptres}.

\section{Discussion}
The simulations and experiments reported here, although performed on a small scale, demonstrate that using a SOM knowledge base to guide the agent's exploratory actions may help achieve a quicker accumulation of higher returns when the target tasks are related to the previously learned tasks. Moreover, the nature of the transfer algorithm is such that even in the case where the source tasks are unrelated to the target task, the learning performance does not exhibit drastic drops, as in the case where value functions of source tasks are directly used to initialize or modify the value function of a target task. Another advantage of the proposed approach is that it can be easily applied to different representation schemes (for example, tabular representations, tile coding, neural networks etc.,), as long as the same action space and representation scheme is used for the target and source tasks. This property has been exhibited in Fig.~\ref{fig:SOMmap}, where SOMs resulting from two different representation schemes are shown.
With regards to the storage of knowledge of learned tasks, the SOM-based approach offers a scalable alternative to explicitly storing the value function weights of all the learned tasks. 
From a practical point of view, one may also define upper limits to the size to which the SOM may expand based on known memory limitations. 

Despite these advantages, several issues remain to be addressed. The most fundamental limitation of this approach is that it is applicable only to situations where tasks differ solely in their reward functions. This may prohibit its use in a number of practical applications. 
Moreover, the approach executes any action advice that it is provided with. The decision to execute the advised actions could be carried out in a more selective manner, perhaps based on the cosine similarity between the target task and the advising node of the SOM. 

One limitation with our approach, as described, is that since the actions are always either greedy or dictated by one of the SOM nodes, every state-action pair is not guaranteed to be visited infinitely often, and hence, $Q$-learning is not guaranteed to converge. However, this issue can simply be addressed by allowing the agent to take random exploratory actions with a very small probability. The final exploration strategy would hence be $\epsilon$-$\beta$-greedy ($\epsilon\ll\beta$), such that with a probability of $\epsilon$, the agent takes random actions, with a probability of $\beta$, it follows the SOM-guided actions, and with a probability of ($1-\epsilon-\beta$), it takes greedy actions. Although we were able to learn good policies in our implementations, a simple modification to the exploration strategy as mentioned above, guarantees the convergence of the $Q$-learning component of our approach.   

Apart from this, and the several other possible variants to this approach, ways to automate the selection of the threshold parameters, establishing theoretical bounds on the learning performance and alternative approaches to quantify the efficiency of the knowledge storage mechanism may be future directions for research. 


\section{Conclusion}
We described an approach to efficiently store and reuse the knowledge of learned tasks using self organizing maps. We applied this approach to an agent in a simulated multi-task navigation environment, and compared its performance to that of an $\epsilon-$greedy approach for different values of the exploration parameter $\epsilon$. Results from the simulations reveal that a modified exploration strategy that exploits the knowledge of previously learned tasks improves the agent's learning performance on related target tasks. Further, navigation experiments were conducted using a physical micro-robotics platform, the results of which validated those obtained in the simulations. 
In addition to being able to leverage previously learned task knowledge for transfer, the proposed approach is also shown to be able to store the knowledge of multiple tasks in a scalable manner. This aspect is demonstrated empirically, and is supported by some analytically derived properties. 
Overall, our results indicate that the proposed approach transfers knowledge across tasks relatively safely, while simultaneously storing relevant task knowledge in a scalable manner. Such an approach could prove to be useful for agents that operate using the reinforcement learning framework, especially for real world applications such as autonomous robots, where scalable knowledge storage and sample efficiency are critical factors.

\section*{Acknowledgements}
This work is supported by the President's graduate fellowship (Ministry of Education, Singapore). 



\end{document}